# TEMPORAL GRAPH NEURAL NETWORKS FOR EARLY ANOMALY DETECTION AND PERFORMANCE PREDICTION VIA PV SYSTEM MONITORING DATA

S. Mukherjee [1,2], L. Vuillon[2], L. Bou Nassif[4], S. Giroux-Julien[5], H. Pabiou[4], D. Dutykh[3], I. Tsanakas[1]

[1]Univ. Grenoble Alpes, CEA, Liten, 73375 Le Bourget du Lac, France
[2]Univ. Savoie Mont Blanc, CNRS, LAMA, Chambéry, 73000, France
[3]Mathematics Department, Khalifa University, Abu Dhabi, PO Box 127788, United Arab Emirates
[4]INSA Lyon, CNRS, CETHIL, UMR5008, Villeurbanne, 69621, France
[5]Université Claude Bernard Lyon 1, CNRS, LAGEPP, UMR5007, Villeurbanne, 69100, France

*Corresponding author: Srijani.Mukherjee@univ-smb.fr   Ph.: +33698703684

ABSTRACT: The rapid growth of solar photovoltaic (PV) systems necessitates advanced methods for performance monitoring and anomaly detection to ensure optimal operation. In this study, we propose a novel approach leveraging Temporal Graph Neural Network (Temporal GNN) to predict solar PV output power and detect anomalies using environmental and operational parameters. The proposed model utilizes graph-based temporal relationships among key PV system parameters, including irradiance, module and ambient temperature to predict electrical power output. This study is based on data collected from an outdoor facility located on a rooftop in Lyon (France) including power measurements from a PV module and meteorological parameters.

The Temporal GNN model integrates Graph Convolutional Networks (GCN) and Gated Recurrent Units (GRU) to capture spatial and temporal dependencies effectively. The model is trained to minimize Mean Squared Error (MSE) loss and is evaluated using Mean Absolute Error (MAE). Anomalies are identified by computing absolute errors and setting a threshold. The proposed framework achieves a MAE of 0.0707 on normalized output power prediction, outperforming traditional methods (e.g., Random Forests, SVMs) and advanced deep learning models (e.g., LSTMs) reported in recent literature, highlighting its potential to set a new benchmark for solar PV performance prediction and anomaly detection. The proposed method holds significant promise for real-world applications, providing actionable insights for maintenance and optimization in solar PV installations.

## 1 INTRODUCTION

The increasing global reliance on solar energy has underscored the critical need for robust and reliable monitoring systems for photovoltaic (PV) installations. Effective performance prediction and timely anomaly detection are paramount to ensuring the long-term efficiency, reliability, and economic viability of these systems. Traditional monitoring methods, often based on simple thresholds or statistical rules, frequently fail to account for the complex interplay of environmental and operational variables that affect PV performance. These methods may lead to high rates of false positives or, more critically, miss subtle but significant anomalies that can indicate underlying system faults.

To overcome these limitations, advanced data-driven approaches are essential. Machine learning and deep learning models have shown promise in this field, offering the ability to learn complex, non-linear relationships from vast datasets. However, many of these models, such as Long Short-Term Memory (LSTM) networks or Support Vector Machines (SVMs), primarily focus on temporal dependencies, treating individual data points as independent sequences. This overlooks the inherent spatial relationships among different system parameters—such as the correlation between irradiance, temperature, and power output—which are crucial for a comprehensive understanding of system behaviour. Recent advances in Graph Neural Networks (GNNs) provide a powerful way to learn from structured data. By representing PV system parameters as graph nodes, and their interdependencies as edges, GNNs can capture both spatial (between parameters) and temporal (across time) correlations.

This study introduces a novel Temporal Graph Neural Network (T-GNN) that integrates Graph Convolutional Networks (GCNs) and Gated Recurrent Units (GRUs) for PV system analysis. Our model is designed to capture both the spatial dependencies between various PV system parameters and their temporal evolution. By representing the PV system as a dynamic graph, where each parameter is a node and its relationships over time are captured by edges, the T-GNN can model intricate dependencies more effectively than traditional methods. The primary objective is to develop a robust framework for accurate power output prediction and early anomaly detection, thereby providing actionable insights for PV system maintenance and optimization. This work represents a significant contribution to the field of PV performance analysis by demonstrating the superior capabilities of a spatiotemporal deep learning model in a real-world application.

## 2 METHODOLOGY

The proposed methodology for PV performance prediction and anomaly detection is a multi-step process centered on the T-GNN model. The workflow begins with data collection and preprocessing, followed by the model architecture design, training, and a two-stage process for prediction and anomaly detection.

## 2.1 DATA COLLECTION AND PREPROCESSING

Real-world monitoring data was collected from a PV system located on a rooftop in Lyon, France (latitude 45.783055° N, longitude 4.873611° E). The system consists of a central PV module (1675x992x35 mm) within a 12° tilted east-west array. Data from 10 sunny days per season (Spring: March-May 2023; Summer: June-August 2023; Autumn: September 2023) was used for this study. The data was recorded at 2-second intervals and includes four key input parameters:

- Global shortwave irradiation (G_sw)
- Global longwave tilted irradiation (G_lw)
- PV module temperature (T_pv)
- Ambient air temperature (T_air)

The target variable for prediction is the electrical power output (P_out). Prior to model training, all input features and the target variable were normalized using MinMax scaling to ensure a consistent range and prevent parameters with larger magnitudes from dominating the learning process. The dataset was then split into an 80-20 training and testing at random.

## 2.2 TEMPORAL GRAPH NEURAL NETWORK ARCHITECTURE

The core of our model is the Temporal GNN, which integrates two primary components to handle both spatial and temporal dependencies. To model the complex dependencies among PV system parameters, we construct a directed temporal graph $G=(V,E)$ where each node $v_i \in V$ represents a system parameter (e.g., irradiance, temperature, power output) at a given time step. Spatial relationships are encoded by directed edges $(v_i^t, v_j^t) \in E$, capturing causal influence between parameters at the same time step, such as irradiance affecting module temperature and temperature influencing power output[1]. Temporal dependencies are modeled by directed edges $(v_i^t, v_i^{t+1})$, linking each node to its future state, thereby preserving the sequential dynamics of the system. The adjacency matrix A of this graph is thus time-indexed, forming a sequence $\{A^t\}_{t=1}^T$. The node feature matrix at time t, denoted $X^t \in \mathbb{R}^{|V| \times d}$, stores the observed values of all parameters. During message passing, node embeddings are updated as

$$h_i^{t+1} = \sigma\left(\sum_{j \in \mathcal{N}(i)} \frac{1}{c_{ij}} W h_j^t + U h_i^t\right)$$

Here, $h_i^{(t+1)}$ represents the hidden state of node *i* at time *t+1*. Where N(i) denotes the neighborhood of node i, W and U are learnable weight matrices, and $c_{ij}$ is a normalization factor. The formula effectively combines a Graph Convolutional Network (GCN) layer for spatial information aggregation and a Gated Recurrent Unit (GRU) for temporal evolution[2]. The GCN layer aggregates information from a node's neighbors, while the GRU processes the sequential data to learn temporal patterns. The full model architecture consists of a Graph Convolution Layer, a GRUCell, and a final Fully Connected Layer (*Fig 1*) to produce the predicted power output, $\hat{P}\ out$. The model was trained using the Mean Squared Error (MSE) loss function with the Adam optimizer, with a learning rate of 0.01 and 100 epochs (*Fig 2*).

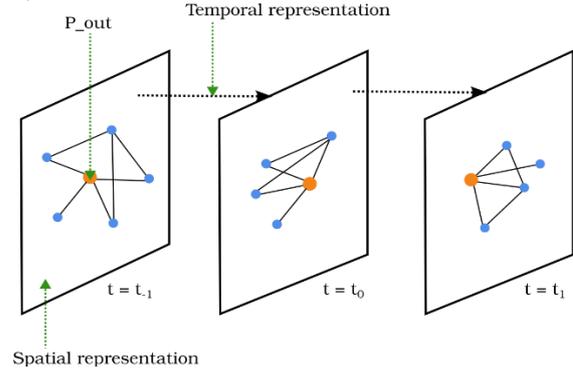

Fig 1: Schematic diagram of the evolution of the spatio-temporal graph prediction

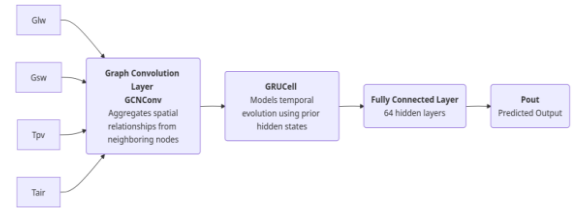

Fig 2: The architecture of the proposed model

## 2.3 ANOMALY DETECTION

Our framework for anomaly detection is a two-step process based on the model's prediction residuals.

Step 1: Power Prediction:
The T-GNN model predicts the power output, $\hat{P}\ out$.

Step 2: Residual Error Calculation:
The residual error, e, is computed as the absolute difference between the actual power output and the predicted power output:

$$e = |P\_out - \hat{P}\_out|$$

Step 3: Z-score Anomaly Detection:

Anomalies are identified using the Z-score of the residuals, which measures how many standard deviations an observation is from the mean of the residuals.

$$Z = \frac{e - \mu\_e}{\sigma\_e}$$

Here, $\mu\_e$ is the mean of the residual errors and $\sigma\_e$ is the standard deviation. A point is flagged as an anomaly if its Z-score magnitude exceeds a predefined threshold, τ. For this study, the anomaly thresholds were set using the Interquartile Range (IQR) method, with a lower bound of $(Q1-1.5*IQR)$ and an upper bound of $(Q3+1.5*IQR)$. Outliers falling outside this range are considered anomalies.

## 3. RESULTS AND DISCUSSION

The T-GNN model demonstrated strong predictive performance on the test set, effectively capturing the diurnal patterns of PV power generation as presented in *Fig 2*.

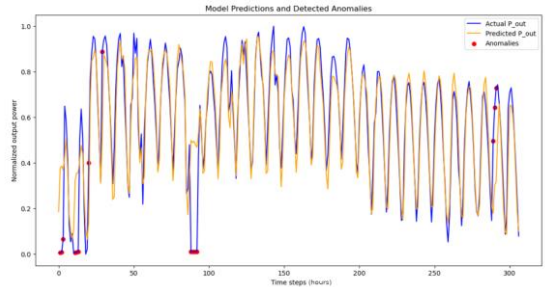

Fig 2

The model achieved a Mean Absolute Error (MAE) of 0.0707 for normalized power output prediction and a Mean Percentage Error (MPE) of 2.26. These results represent a significant improvement over existing state-of-the-art methods[3]. A comparison of our model's performance with other methods is summarized in *Table 1*.

| Model | Notes | MAE |
|---|---|---|
| T-GNN (ours) | Spatio-temporal | 0.0707 |
| LSTM | Temporal only | 0.09–0.10 |
| RF / SVM | Traditional ML | 0.10–0.15 |

Table 1: Comparison of Model Performance by Mean Absolute Error (MAE)

The T-GNN's superior performance can be attributed to its unique ability to learn from both the spatial relationships among input parameters and their temporal evolution, a capability that traditional models lack. As shown in the figure comparing actual and predicted power output, the model's predictions closely track the actual data, even during periods of rapid change.

The anomaly detection component of the model successfully identified some inconsistent power readings, which constituted 5.21% of the dataset. These anomalies were flagged as points where the model's prediction significantly deviated from the actual measured power output. A box plot in *Fig 3* of the absolute errors with anomaly thresholds clearly visualizes these outliers.

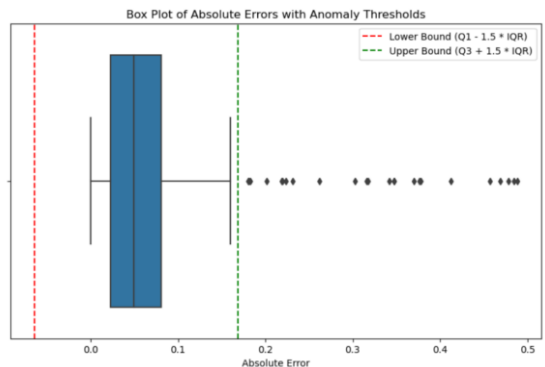

Fig 3

Cross-validation of the detected anomalies confirmed that they corresponded to genuine inconsistencies in the measured data, such as sudden and unexplainable drops in power output, which may indicate a measurement fault or an operational irregularity. The low percentage of detected anomalies suggests that the model is highly selective and robust[4], minimizing the risk of false positives. This makes the framework a valuable tool for real-time monitoring, as it can alert operators to potential issues without generating excessive noise.

## 4. CONCLUSION AND FUTURE RESEARCH

This study successfully demonstrates the effectiveness of a Temporal Graph Neural Network for both predicting PV power output and detecting anomalies in real-time. By leveraging a T-GNN, our model can effectively capture the complex spatiotemporal dependencies inherent in PV system data, leading to significantly higher predictive accuracy compared to traditional machine learning and deep learning methods. The low MAE and MPE achieved by our model, combined with its ability to precisely identify anomalies, positions it as a promising solution for enhancing the reliability and efficiency of solar PV systems.

The insights gained from this work provide a clear path toward developing more intelligent and resilient PV monitoring systems. Future work will focus on expanding the model's capabilities by incorporating additional data sources, such as electrical parameters (e.g., voltage and current) and infrared (IR) thermal images. This multi-modal approach is expected to further enhance the model's ability to detect more subtle and complex types of anomalies, providing a more comprehensive diagnostic tool for PV system maintenance and optimization.